# All models are local: time to replace external validation with recurring local validation

Alex Youssef[1,2], Michael Pencina[3], Anshul Thakur[2], Tingting Zhu[2], David Clifton[2,4], Nigam H. Shah[5-7]

**Author affiliations:**
[1]Stanford Bioengineering Department, Stanford University, Stanford, CA, USA
[2]Department of Engineering Science, University of Oxford, Oxford, UK
[3]Duke University School of Medicine, Durham, NC, USA
[4]Oxford-Suzhou Centre for Advanced Research, Suzhou, China
[5]Center for Biomedical Informatics Research, Stanford University School of Medicine, Stanford, CA, USA
[6]Technology and Digital Solutions, Stanford Medicine, Stanford, CA, USA
[7]Clinical Excellence Research Center, Stanford Medicine, Stanford, CA, USA

**Corresponding author:**
alexeyyoussef@alumni.stanford.edu

**Abstract:**
External validation is often recommended to ensure the generalizability of ML models. However, it neither guarantees generalizability nor equates to a model's clinical usefulness (the ultimate goal of any clinical decision-support tool). External validation is misaligned with current healthcare ML needs. First, patient data changes across time, geography, and facilities; these changes create significant volatility in the performance of a single fixed model (especially for deep learning models, which dominate clinical ML). Second, newer ML techniques, current market forces, and updated regulatory frameworks are enabling frequent updating and monitoring of individual deployed model instances. We submit that external validation is insufficient to establish ML models' safety or utility. Proposals to "fix" the external validation paradigm do not go far enough. Continued reliance on it as the ultimate test is likely to lead us astray. We propose the MLOps-inspired paradigm of *recurring local validation* as an alternative that ensures the validity of models while protecting against performance-disruptive data variability. This paradigm relies on "site-specific" reliability tests before every deployment, followed by regular and recurrent checks throughout the life cycle of the deployed algorithm. Initial and recurrent reliability tests protect against performance-disruptive distribution shifts, and concept drifts that jeopardize patient safety.

**Introduction**
Historically, predictive models followed a standard development pipeline: model development & internal validation, external validation, and clinical impact studies [1]. During internal validation, model performance is evaluated against the same dataset used for development [2]. During external validation, the model is tested against single or multiple datasets representing populations different from the training population [1]. A special case of external validation is temporal validation, where the model is evaluated based on data collected from the training population but at a different time [3]. Internal-external validation studies perform external validation during the initial development stage [4,5] or after development by the same or an outside team [6]. Ideally, validation studies should be followed by real-world studies evaluating the deployed models' clinical impact and usefulness [7]. However, such studies are rarely performed [8,9]. Frequently, external validation ends up being the de-facto test to evaluate ML models before deployment.

**External validation in healthcare ML**
External validation originated as a tool to measure generalizability: the cross-population replicability of performance and transportability of a model [10]. Generalizability is used as a proxy for reliability: the stability of the discrimination and calibration across time, deployment locales, and target populations [11]. In the medical literature, external validation is often considered the ultimate test to conclusively judge an ML model's safety, reliability, and generalizability [1,6,8,9,12]. According to this logic, if a model passes an external validation test on one or a few datasets, it is generalizable, safe, and reliable. However, external validation does not guarantee generalizability [13]. Neither does it equate to model usefulness, which should be the true goal of any clinical decision-support tool. Taken together, we question whether external validation should be a gold standard paradigm for evaluating healthcare ML algorithms.

**External validation fails to generalize**
Ideally, the datasets used in external validation reflect the populations in whom the final use of the model is intended. However, these datasets are frequently chosen based on convenience and availability. As such, datasets used for external validation usually do not represent the populations of intended use. Ramspek and colleagues [2] propose performing external validation tests when transporting a model across two different care settings (e.g., primary to secondary care). There is growing recognition that we cannot rely on external validation to make broad and deterministic claims of generalizability across unseen data because data heterogeneity cannot be captured by a single or a few external validation datasets [2,8,14]. Cabitza and colleagues demonstrate that a model can perform and generalize well to some datasets but not others [10]. They introduce meta-validation: a validation technique that measures data similarity and cardinality to quantify the difference between the external validation dataset and the original training dataset [10]. This is a step in the right direction. Such approaches seek to quantify differences between training and implementation datasets and recommend repeating external validation tests before implementing the model on a population perceived to be "different enough" from that of development. However, such approaches still assume that external validation on a few datasets can reliably determine a model's performance on other datasets.

**External validation does not equate to clinical usefulness**
External validation studies usually rely on global performance statistics (e.g., AUROC, AUPRC) that do not capture the practical value of the clinical decision support (CDS) tool [15]. Clinical utility and safety are measured by metrics beyond the replicability of a narrow performance statistic across settings. Clinically useful models recommend decisions that improve outcomes or reduce cost [7,16–25]. Fair models achieve parity in the statistical outcome/model recommendation across various population groups [26–28]. Simply put, the metrics that are critical for safety and clinical utility rely on site-specific, guideline-based, translation of the ML algorithm into clinical recommendations [29]. These are hard to adequately capture in a generic external validation.

**External validation is misaligned with how ML models are built, shared and sold**
In practice, an external validation exercise only supports reliability and generalizability claims limited to the specific population in the specific healthcare facility during a particular time window. This places external validation at odds with how ML models in healthcare are built, shared, and sold. It is infeasible to assume we can always identify *"target populations"* and validate models on representative datasets before implementation.

The reality is that ML solutions will be developed and brought to the market by commercial entities seeking to implement and sell their models across a wide range of geographies and populations (e.g., facilities across the US and Europe). Epic is one of the leaders in offering EHR-integrated predictive analytic tools with multiple commercially available models. The firm released its first Epic Sepsis prediction Model (ESM) as a one-size-fits-all solution reporting an AUC performance threshold of 0.73-0.83 from a calibration performed on three healthcare systems from 2013 to 2015 [30]. Later studies reported a significantly lower AUC performance of 0.60-0.64 (slightly better than chance) [31,32], raising concerns about potential patient harm. This triggered waves of criticism and calls for local validation and fine-tuning of predictive models [33]. Epic recently and appropriately announced that it would localize the new version of ESM to a given site's patient-mix. Thus abandoning the notion of a generalizable "one-size-fits-all" model, which the research community had argued for in the past, and embracing a site-specific localization and validation strategy [34].

External validation is particularly problematic with Deep Learning (DL) models, given their sensitivity to distribution shifts and data variability. The current state-of-the-art DL solutions tend to overfit the training distribution and often fail to generalize on out-of-distribution test data [35–37]. In other words, the deviation in training and test populations can significantly affect the performance of DL models. This places DL models at odds with external validation as an evaluation paradigm because it strives for universal generalizability.

In addition, distribution shifts make the expectation of universal generalizability a problematic notion in healthcare. When the model inputs are not purely biological, and include

"operational inputs" (i.e., those that are about the nature of care delivery), then generalizability is impossible to guarantee across time, geography, and facilities [34]. For example, Yu and colleagues demonstrate that almost all models have a decline in performance when tested on a dataset outside the distribution of the training dataset [8]. Singh and colleagues tested a mortality prediction model on 179 healthcare facilities across the US, showing that distribution shifts are a universal phenomenon across facilities and that these shifts significantly impact the model's performance [12].

An ML/DL model that includes "operational inputs" will not (and perhaps should not!) generalize to all populations and healthcare facilities. In fact, if a model – such as a readmissions predictor – worked equally well across locales such as Palo Alto, Durham, and Mumbai; one has to question how that is possible given the dramatically different patient-mixes, care-protocols, and data collection processes. We conclude that current approaches that use external validation to make deterministic and broad conclusions about generalizability and subsequent reliability can lead us astray.

**Our recommendation**
It is a fallacy to judge a model's generalizability, reliability, safety, or utility from external validation alone – especially when operational inputs are in use.  While external validation helps build confidence in a model's performance on a well-defined population (e.g., the population of a particular hospital catchment area during a particular time window), many of today's models are intended for broad geographic implementation markets and many different and diverse implementation populations.  Taken together, we suggest that a better use of the essence of external validation would be "site-specific" validation performed as a reliability test before every local deployment and repeated on a recurring basis. It builds on and expands the concept of temporal validation. Such local validation would be performed (1) before deployment at a particular facility, given the novelty of the unseen local dataset, and (2) repeated over time, given the potential for performance-disruptive distribution shifts and concept drifts.

This "recurring local validation" paradigm is new to healthcare but standard practice in Machine Learning Operations (MLOps). MLOps is a discipline concerned with the at-scale production, deployment, monitoring, and maintenance of models. MLOps originates from a 2015 paper by Google [38]. MLOps frameworks are being widely adopted in industrial ML deployment. Continuous performance monitoring and model updating are critical for ML implementation through MLOps. Practically, this includes continuously monitoring the performance of the local model instances and updating them to maintain the desired level of model performance [39].

In principle, a "recurring local validation" paradigm would rely on the presence of historical data to perform the initial pre-deployment reliability tests. It is reasonable to expect such access to become the norm in healthcare. Initial model validation can be followed by implementing a default configuration model in "silent mode": model output is recorded and evaluated against the clinical ground truth to assess local performance [40]. The model is then fine-tuned using the collected data during the silent phase. Even small amounts of local data collected during a short time-frame can be valuable to localize a model. The "silent mode" approach can also be adopted when historical data is not available. A recent study has demonstrated that a non-disease-specific deterioration prediction model can be fine-tuned to reliably predict deterioration in COVID-19 patients with small amounts of COVID-19 data collected in the first few weeks of the pandemic [41].

Still, one can argue that in instances of data unavailability, a model developed and externally validated on a sufficiently similar population could be suitable for immediate deployment in such a scenario. Such assumptions run into the impracticalities of measuring similarity/difference between populations (e.g., we do not have validated measures of population similarity), and the fact that two facilities may have a similar patient-mix but vastly different care workflows, processes, and equipment that may disrupt the performance of a model. Therefore, a pilot phase as part of a "recurring local validation" paradigm would still be appropriate.

While reliability is a necessary goal for healthcare ML developers, aiming for universally generalizable models evaluated through external validation is a flawed way to achieve reliability. Instead, the "recurring local validation" via MLOps is a well-traveled path to create reliable models through retraining and fine-tuning, incremental learning, and continual learning. Such frameworks leverage the dynamic adaptive nature of AI algorithms. Model architectures, hyperparameters, and weights can be adapted at various deployments and over time, preserving reliability while protecting performance against data shifts and concept drifts [39,42–44].

**Conclusion**
The healthcare ML community is often led to view external validation as a gold-standard evaluation. Yet, the external validation paradigm does not ensure the generalizability of a model across time, populations, and geographies. Neither does it provide insights into a model's fairness and clinical usefulness. Proposals to "fix" external validation go in the right direction but do not go far enough. We show that external validation is misaligned with how models in healthcare are built, shared, and sold. We submit that external validation is insufficient and may not even be necessary or critical to establish the reliability or utility of ML models – especially for models that rely on operational inputs. We propose the MLOps-inspired paradigm of "recurring local validation" to maintain the validity of models while protecting against performance-disruptive data variability. We should routinely and continuously perform local evaluations of models that guide care.

**Conflicts of interest:**
The authors declare no competing conflicts of interest.

**Funding statements:**
DAC is funded by an RAEng Research Chair and an NIHR Research Professorship, in addition to support from the NIHR Oxford Biomedical Research Centre, the InnoHK Centre for

Cerebro-cardiovascular Engineering, the Oxford Pandemic Sciences Institute, and the Oxford-Suzhou Centre for Advanced Research (China). TZ is supported by the Engineering for Development Research Fellowship provided by the Royal Academy of Engineering. The views expressed are those of the authors and not necessarily those of the NHS, the NIHR, the Department of Health, or InnoHK – ITC.